\documentclass[a4paper]{IEEEtran}
\pdfoutput=1
\IEEEoverridecommandlockouts
\def\IEEEtitletopspace{16pt}
\usepackage{amsmath,amssymb,amsfonts}
\usepackage[numbers,sort&compress]{natbib}
\usepackage{algorithmic}
\usepackage{graphicx}
\usepackage{textcomp}
\usepackage[table]{xcolor}
\usepackage{subcaption}
\usepackage{caption}
\usepackage{mathptmx}
\usepackage{gensymb}
\usepackage{courier}
\usepackage[paper=a4paper,top=104pt, left=55pt, right=34pt, bottom=57pt]{geometry}

\usepackage{hyperref}
\usepackage[capitalise]{cleveref}
\usepackage[scaled=0.9]{helvet}

\usepackage{enumitem}
\usepackage{multicol}
\usepackage{booktabs}

\begin{document}
\pagenumbering{gobble}
\title{Camber-changing flapping hydrofoils for efficient and environmental-safe water propulsion system\\
\thanks{
$^{1}$TUM eAviation group, Department of Aerospace and Geodesy
\par $^{2}$Aerial Robotics Laboratory, Imperial College London
\par $^{3}$Laboratory of Sustainability Robotics, EMPA, Dübendorf, Switzerland.
\par $^{4}$EPFL, Switzerland.}
}
\author{Luca Romanello$^{1,2}$, Leonard Hohaus$^{1}$, David-Marian Schmitt$^{1}$, Mirko Kovac$^{3,4}$ and Sophie F. Armanini$^{2}$ 
 }
\maketitle
\begin{abstract}
    This research introduces a novel hydrofoil-based propulsion framework for unmanned aquatic robots, inspired by the undulating locomotion observed in select aquatic species. The proposed system incorporates a camber-modulating mechanism to enhance hydrofoil propulsive force generation and eventually efficiency. Through dynamic simulations, we validate the effectiveness of the camber-adjusting hydrofoil compared to a symmetric counterpart. The results demonstrate a significant improvement in horizontal thrust, emphasizing the potential of the cambering approach to enhance propulsive performance. Additionally, a prototype flipper design is presented, featuring individual control of heave and pitch motions, as well as a camber-adjustment mechanism. The integrated system not only provides efficient water-based propulsion but also offers the capacity for generating vertical forces during take-off maneuvers for seaplanes. The design is tailored to harness wave energy, contributing to the exploration of alternative energy resources. This work advances the understanding of bionic oscillatory principles for aquatic robots and provides a foundation for future developments in environmentally safe and agile underwater exploration.
\end{abstract}



\section{Introduction} \label{chap:intro}
\begin{figure}[htbp]
    \centering
    \includegraphics[width=0.9\linewidth]{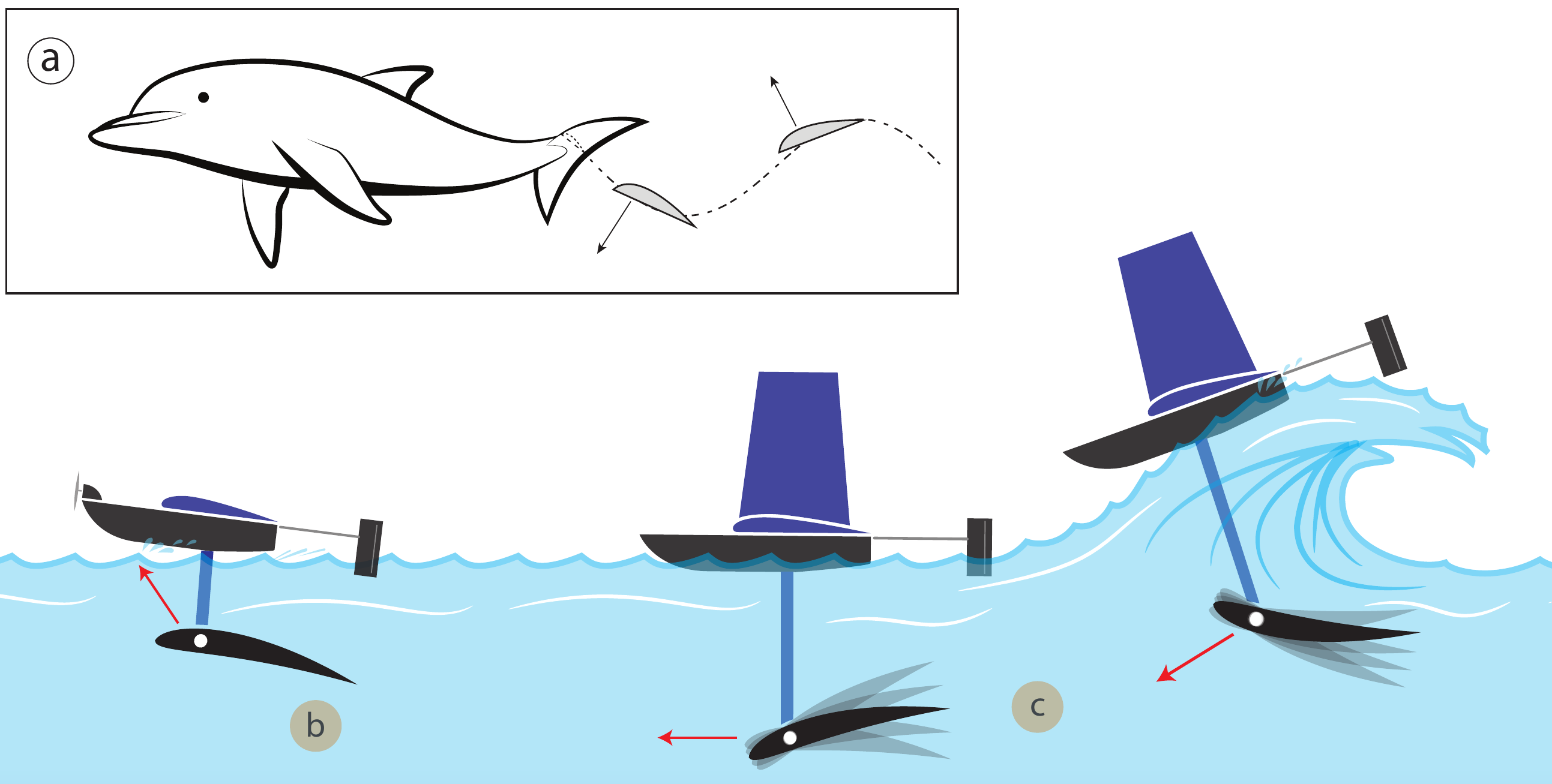}
    \caption{Working principles of the system. (a) Swimming maneuvers in dolphins in water resembling hydrofoil}
    \label{fig:working_principles}
\end{figure}

Unmanned aquatic robots, devised for surface-level exploration, are assuming increasing significance in driving forward research and environmental surveillance in marine and freshwater domains \cite{saildrone,smurf}. Water environmental sensing requires vehicles that are quiet to avoid disturbing data collection and aquatic species, while also being environmentally safe for both air and water inhabitant species.  
Such robots incorporate diverse propulsion mechanisms for aquatic navigation, encompassing jet propulsion \cite{Lin2012DevelopmentOA}, water propellers \cite{6093749}, wave gliding \cite{wave_glider}, and bionic fin and foot designs \cite{cormorant, beaver} as aquatic propulsion methods. Additionally, air-based propulsion via sails or redundant systems like aerial propellers, as for aerial-aquatic vehicles and seaplanes can be also utilized \cite{seaplane, sailmav}. 
A bio-mimetic fins and feet setup characterized by a slower motion exhibits a heightened degree of environmental safety when contrasted against high-impact propulsion modalities like aquatic propellers or jet propulsion systems. Moreover, this approach is not dependent on meteorological and aquatic parameters, as for the case of wave gliders or sail-based configurations. 
Drawing inspiration from aquatic organisms, a concept akin to the fin system has emerged  in the form of a morphing hydrofoil-based low-frequency propulsion framework as a viable propulsion approach for underwater robots. This system aims to replicate the undulating locomotion observed in select aquatic species, as depicted in \cref{fig:working_principles}a. This locomotion comprises a biphasic sequence involving a downward stroke (downstroke) succeeded by an upward stroke (upstroke). Oscillatory swimming relies on the principle of forcing a fin or fluke to move perpendicular to the direction of propulsion, thereby creating lift forces with forward components that becomes known as thrust.
To further enhance the proposed propulsion method, we introduce a camber-modulating mechanism. The primary aim of this system is to enhance the efficiency and propulsive force generation. Introducing appropriate camber adjustments leads to heightened lift-to-drag ratios and lift coefficients compared to a symmetrical foil case. Therefore, we postulate that a hydrofoil that can camber in the right direction will improve the propulsive performance of the bionic oscillatory principle in both upstroke and downstroke motion. In this paper we provide calculations that support this claim and discuss the design of the physical and control structure of a prototype that can be used to test the proposed propulsion principle.
Our approach stands in contrast to conventional fin-type propulsion systems, which rely on drag-induced propulsion \cite{jmse10060783} and exhibit comparatively lower levels of efficiency, as the thrust is produced with lift generation.
Moreover, the implementation of the system on a aquatic vehicle, not only furnishes a water-based propulsion system but also furnishes the capacity of allowing for shorter take-off distances or reduced power consumption for aerial vehicles and seaplanes.
As discussed in \cref{chap:simulation}, varying the camber also allows for inversion of the lift forces, as shown in \cref{fig:working_principles}b, which can be exploited to take-off of aerial-aquatic vehicles.
Furthermore, the system design is tailored to harness the latent energy within waves, inspired by the principles of wave gliders. This approach effectively utilizes wave power for forward propulsion, accessing a substantial and influential energy resource.
In this work we simulate and compare the dynamic behavior of a flipper equipped with a symmetric hydrofoil and with the proposed camber-changing one 
\cref{chap:simulation}. To demonstrate the proposed propulsion concept we then design, manufacture a physical prototype \cref{chap:prototype}. 
 
\section{Concept, Modeling and Simulation} \label{chap:simulation}

The notion of a bionic flapping hydrofoil has garnered substantial and thorough exploration within academia, particularly through the endeavors of Triantafyllou's team 
\cite{triantafyllou1993optimal,triantafyllou1995efficient,anderson1998oscillating}.
Drawing inspiration from their work in studying symmetric hydrofoils' performance, this study seeks to extend the concept by employing a hydrofoil which changes the cambering side based on the direction of the stroke movement, aiming to enhance its performance.
To validate the camber-changing approach, we have conceptualized a flipper mechanism housing two distinct actuated hydrofoils, each endowed with an adaptable camber-adjustment mechanism. This design configuration allows for differential thrust and steering capabilities via the hydrofoil mechanism providing the robot with enhanced agility and maneuverability. One approach to address the challenge of computing the system's dynamics involves an examination of the hydrofoil's motion.

\subsection{Study of motion}
Our focus lies in investigating motion and dynamics to quantify forces and assess the impact of camber variation on these forces. 
\begin{figure}[h] 
  \centering
  \includegraphics[width=0.95\linewidth]{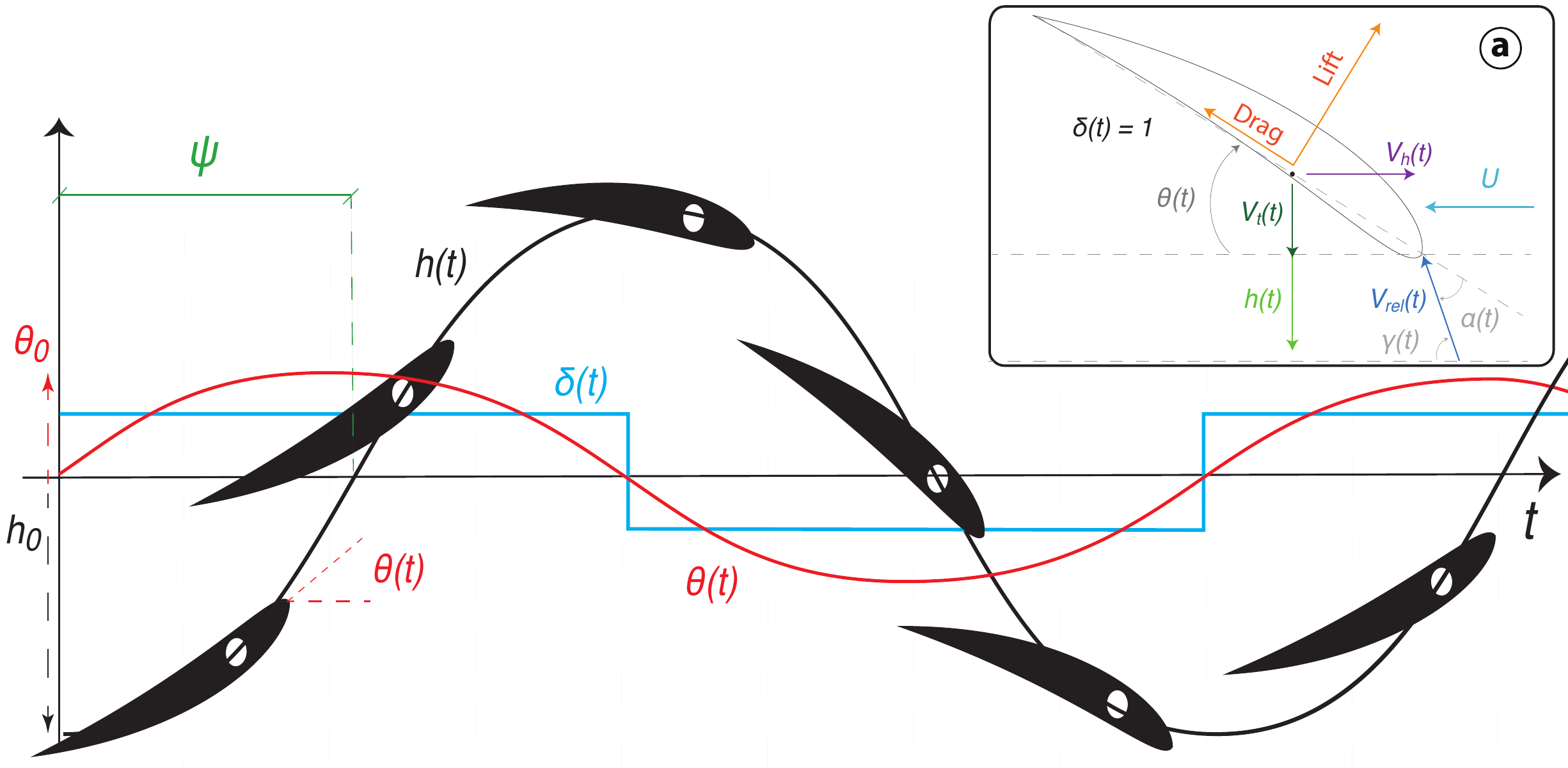}\\
  \caption[]{The complete temporal cycle of the flipper's heave (in black), pitch (in red), and camber (in light blue). (a) Relative velocities of flow and the influential angles of a hydrofoil cross-section at an instant during the periodic oscillation.} 
  \label{fig:kinematics}
\end{figure} 
In the scenario of an undisturbed water flow characterized by a consistent absence of waves or currents, the representation of water motion necessitates the utilization of the relative velocity flow denoted as $V_{rel}$. This vector quantity comprises two components: the forward velocity $V_{h}$, possessing equal intensity but opposite direction to the water velocity flow $U$, and the vertical velocity $V_{t}$ associated with the hydrofoil's heave motion. The latter corresponds to the temporal rate of change, i.e., derivative, of the heave motion. Finally, $\gamma(t)$ signifies the angle of the water relative velocity concerning the water flow direction.
\begin{subequations}
\begin{gather}
V_{rel}(t) = \begin{vmatrix} V_h(t) \\V_t(t)\end{vmatrix}; \quad
V_t(t) = \frac{d}{dt}h(t) \\
\gamma(t) = \arctan{\frac{V_t(t)}{V_h(t)}} \quad \alpha(t) = \gamma(t) - \theta(t)
\end{gather}
\label{ch3:kinematics}
\end{subequations}
The angle $\alpha(t)$ represents the angle of attack, between the relative velocity $V_{rel}(t)$ and the hydrofoil chord. 
The mathematical characterization of the hydrofoil motion involves equations describing heave motion ($h(t)$), pitching motion ($\theta(t)$), and cambering ($\delta(t)$). The complete temporal cycle of the mechanism is illustrated in \cref{fig:kinematics}, where pitching is highlighted in red and heaving is depicted in black. The plot also displays the temporal variation of camber $\delta(t)$, shown in light blue. 
The kinematic behaviors are mathematically described as follows:
\begin{subequations}
\begin{gather}
h(t) = h_0sin(2\pi f t - \Psi)  \\
\theta (t) = \theta_{0} sin(2\pi f t) \quad
\delta (t) = sign(\theta (t)) 
\label{ch3:eq:kinematics}
\end{gather}
\end{subequations}
Pitching and heaving follow sinusoidal patterns characterized by specific parameters: a frequency denoted as $f$, a maximum heave amplitude represented by $h_0$, and a maximum pitching angle denoted as $\theta_0$. Additionally, a phase shift $\Psi$ exists between the heaving and pitching motions. The cambering $\delta(t)$ showcases a cyclic pattern, alternating its sign periodically, in sync with the pitching motion direction. 

\subsection{Dynamics in Simulation}
The computation of the dynamics model relies exclusively on the incoming relative flow interacting with the hydrofoil, characterized by a specific angle of attack at any time point. It is crucial at this point to list the main assumptions we have been adopted. 
These include utilizing constant hydrodynamic properties in the form of lift and drag coefficients. These coefficients were derived from datasets generated using the airfoil design tool XFOIL\cite{airfoil}. Other assumptions include an idealized cambering shape free from physical imperfections, and instantaneous camber adjustments without any latency.

Moreover, the two hydrofoils have been evaluated in isolation, neglecting drag or other interference effects due to the flipper. 
The hydrofoil under consideration transitions from a symmetric NACA 0012 profile, utilized in \cite{pedro}, to an asymmetric NACA 6412 profile. This design is particularly suited for underwater motion characterized by low Reynolds numbers. 
\begin{center}
\begin{tabular}{|c|c|c|c|c|c|c|c|} 
 \hline
\rowcolor{lightgray} $V_h$ & $St$ & $c$ & $s$ & $h_0 / c$ & $\theta_0$ & $\alpha_{max}$ & $\Psi$    \\ [0.5ex] 
 \hline
 0.1 & 0.3 & 0.1 & 0.1 & 0.75 & $10^{\circ}$ & $10^{\circ}$ & pi/2 \\ 
 \hline
\end{tabular}
\end{center}
The computation of the dynamics entails initiating the process by defining the motion frequency (\textit{f}). Pertinent parameters for consideration encompass the vehicle's forward velocity (\textit{$V_h$} in m/s), Strouhal number ($St$), hydrofoil chord length and span ($c$ and $s$), ratio of heave amplitude to chord ($h_0/c$), maximum pitching amplitude ($\theta_0$ in degrees), phase difference ($\Psi$ in radians), and angle of attack ($\alpha(t)$). The frequency is characterized with the formulation $f = \frac{U St}{2 h_0}$.
The equations \cref{ch3:thrust} allow to define the Thrust Force $F_{th}$ and Vertical Force $F_{v}$, which correspond respectively to the forward and upward contributions of the lift and drag forces.  
\begin{subequations}
\label{ch3:thrust}
\begin{gather}
[F_{l}, F_{d}]  = \frac{\rho||V_t,U||^2 c s}{2} [C_l(\alpha(t)), C_d(\alpha(t))]\\
\begin{vmatrix} F_{th} \\F_{v}\end{vmatrix}  =  \begin{vmatrix} \sin(\theta(t)) \quad - \cos(\theta(t)) \\ \cos(\theta(t)) \quad  \sin(\theta(t)) \end{vmatrix} \cdot \begin{vmatrix} F_{l} \\F_d\end{vmatrix}
\end{gather}
\end{subequations}

The Strouhal numbers is also consistent with the same under effective propulsion of birds and fish in nature \cite{taylor2003flying}.

\begin{figure}[h] 
  \centering
  \includegraphics[width=\linewidth]{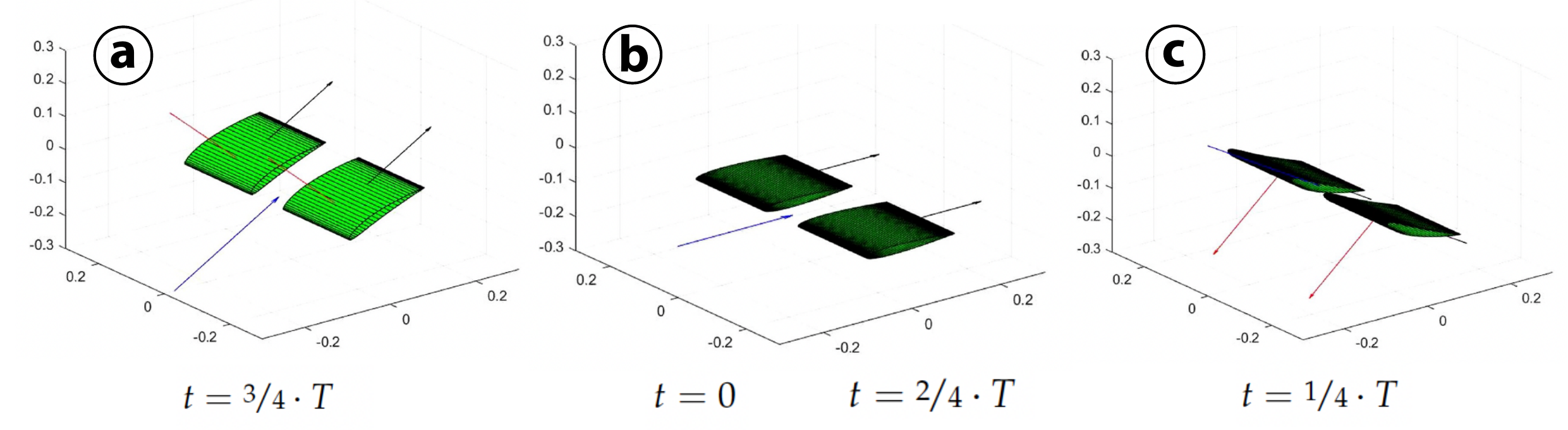}\\
  \caption[]{Lift (red) and drag (black) forces simulation for the downstroke motion (a) intermediate and straight position (b) and upstroke motion (c)} 
  \label{fig:simulation_forces}
\end{figure}

The simulation assumes that the system has already attained a velocity of 0.1 m/s, corresponding to a water flow velocity ($U$) at the same magnitude, under the assumption of a steady water condition. 
The simulation calculates forces over a single oscillatory cycle, defining the mechanism's motion based on parameters from \cite{anderson1998oscillating} for optimal efficiency in their experiments. 
\cref{fig:simulation_forces} depicts the vectors that represent the incident flow (blue), the direction of lift (red), and the direction of drag (black), relative to the cambering hydrofoils' motion, all of which are normalized.

The simulation results, encompassing variations in Horizontal Thrust and Vertical forces throughout the complete cycle, are depicted in Figure \ref{fig:simulation_forces} for both the cambering and symmetric solutions. The data reveals a notable enhancement of forward thrust with the implementation of the cambering solution, resulting in a 40\% increase compared to the symmetric solution. This finding underscores the potential of the cambering approach to substantially improve the propulsive performance of the investigated system. As expected, the peak of the vertical forces occurs around null pitching angle, with the lift force pointing upward, representing a beneficial force aiding in seaplane takeoff. Nevertheless, when examining the magnitude of the vertical force throughout the cycle, there is a potential drawback as it introduces the risk of unwanted system oscillations.

\section{Design and System} \label{chap:prototype}
\begin{figure}[h]
  \centering
  \includegraphics[width=0.8\linewidth]{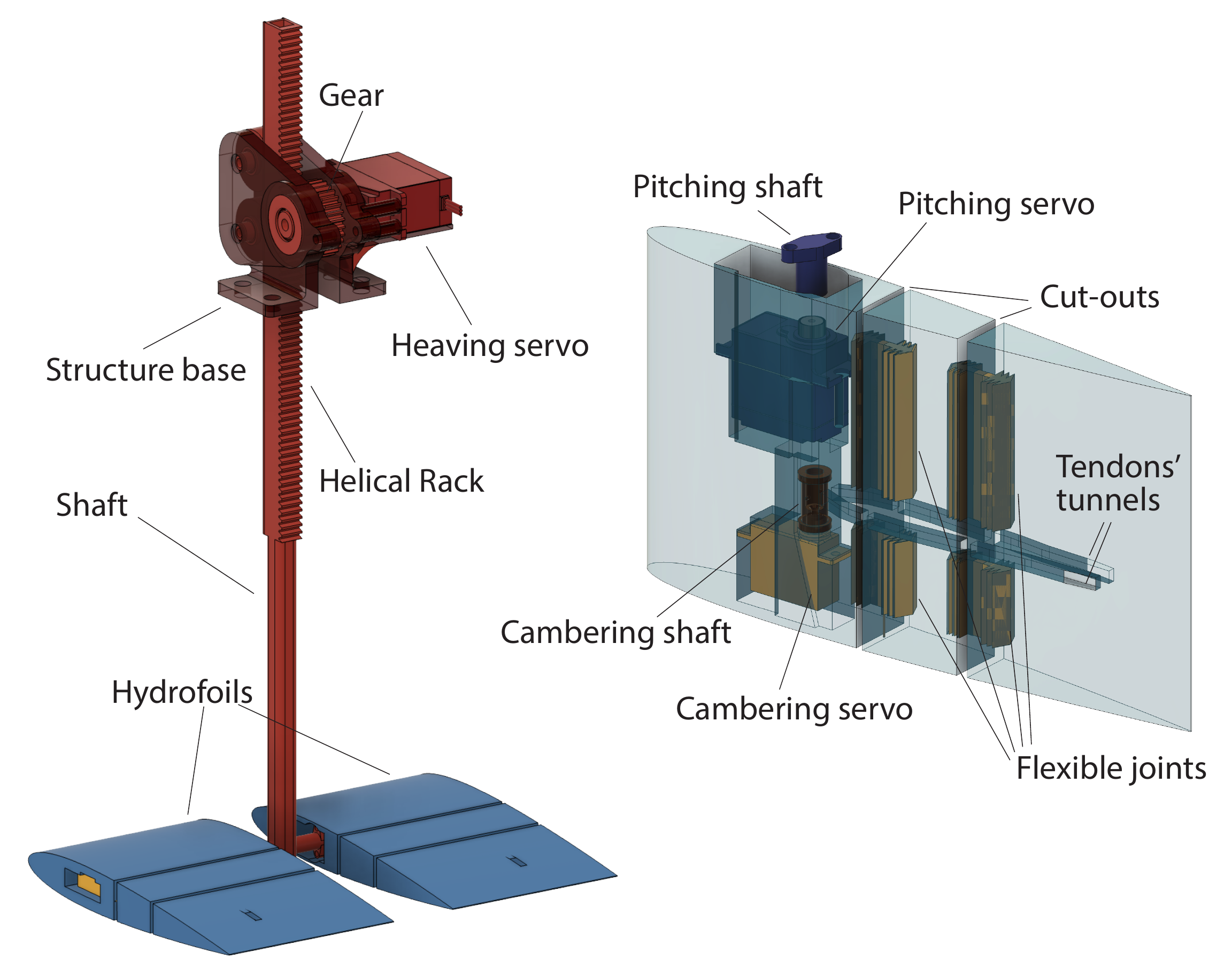}\\
  \caption[]{Flipper CAD model (left) and focus on the camber-changing hydrofoil (right)} 
  \label{fig:control_feet}
\end{figure}

To validate the hypothesis that a camber-adjusting hydrofoil system offers greater propulsive force compared to a symmetric counterpart, a dual hydrofoil flipper has been developed. This flipper is designed for perpendicular heaving motion relative to forward movement. 
With reference to \cref{fig:control_feet}, the fin-like structure exhibits a pivotal pitching rotation of the hydrofoil around its aerodynamic center, further accompanied by a vertical heave motion perpendicular to the water flow direction. 
\begin{figure}[h]
  \centering
  \includegraphics[width=0.8\linewidth]{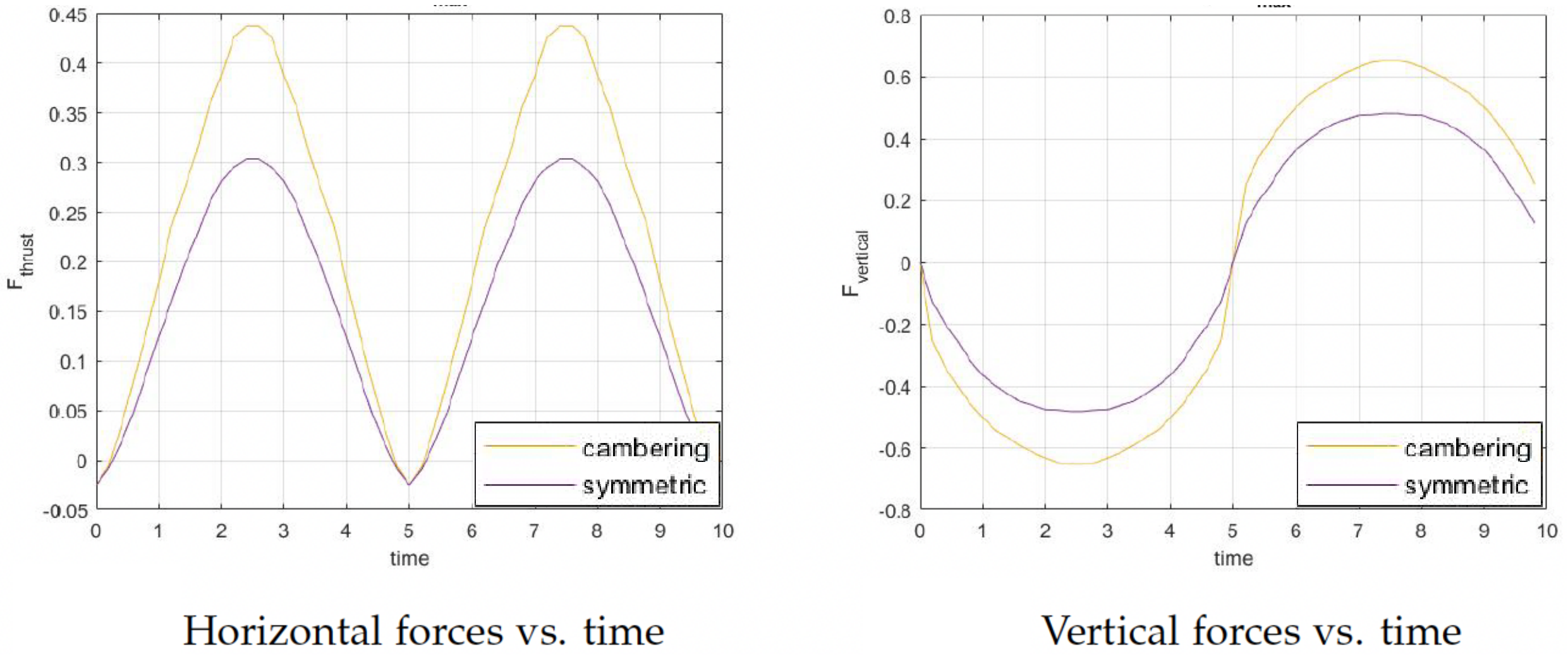}\\
  \caption[]{Vertical and horizontal forces plot in one cycle} 
\end{figure}
The heaving mechanism includes a helical raft and gear setup connected to a continuous motion servo motor. This design prioritizes traits such as lightweight construction, waterproofing, environmental safety, and the ability to harness wave energy.
Regarding the camber-adjustment mechanism, a design incorporating cut-outs and flexible joints enables transition from symmetric to cambered hydrofoil configurations on either side, as shown in \cref{fig:control_feet}. 

\subsection{Design of the Flipper}

The flipper design adheres to the same geometry specifications used in our simulations, notably accommodating a $\frac{h_0}{c}$ ratio ranging from 0.5 to 1 to size the shaft. Using a hydrofoil chord of 0.1m yields a maximum vertical motion range $h_0$ of 0.1m. The heave amplitude, relative to the entire downstroke and upstroke motion, equals twice this amplitude, defining the required space for the heaving mechanism. Achieving this span involves a flipper design that incorporates a helical rack of the specified length, coupled with a gear system converting motor rotation to linear displacement.

The gear sizing must consider the operational frequency, which can be adjusted to test the two systems under different conditions. The flipper is designed to function at frequencies up to 0.5 Hz, corresponding to a maximum velocity of 3 m/s, assuming $St$ = 0.3. Gear sizing is also influenced by the servo's rotation frequency limitation.

\subsection{Design of the Cambering system}
To facilitate tests involving the camber-changing hydrofoil, the design incorporates cut-outs that allow for controlled deformations, thereby enabling alterations to its shape. The strategy is to approximate a NACA6412 foil by manipulating the cutouts on both sides of the NACA0012 foil’s cross-section. This design feature is depicted in Figure \ref{fig:control_feet}. The camber adjustment is achieved through a tendon-driven system, which involves the manipulation of a wire to pull the hydrofoil's tip in different directions, consequently altering its shape and camber.


An essential part of the design is to faithfully replicate the NACA6412 foil shape while accommodating the servos. Referring to \cref{fig:control_feet}, the design layout demonstrates that due to the necessity of accommodating servos for both pitch and camber adjustments, the actual cut-outs implemented for the camber-changing hydrofoil were limited to three. This constraint results in an inability to achieve the the exact NACA 6412 shape, potentially leading to a performance reduction. Moreover, The introduction of cut-outs in the foil surface causes turbulence, especially during bending, leading to increased resistance. Notably, the three distinct sections are interconnected by joints composed of flexible materials.
The camber-modification action is executed through a servo linked to a shaft that controls the unwinding of tendons in both directions. These tendons traverse dedicated tunnels and are anchored at the hydrofoil's extremity.
Conversely, the pitching motion is controlled by a servo integrated into the hydrofoil structure, allowing rotation around the flipper shaft positioned at the hydrodynamic midpoint, known as the 1/3 chord point. Each hydrofoil has its own servo mechanism for individual thrust control. This enables independent pitching motions, facilitating differential thrust capability for generating varying thrust forces.
The camber-changing feature introduces a redundant control for managing differential thrust. By separately modulating the cambering of each hydrofoils it is possible to produce different thrust facilitating the steering maneuvers.

In conclusion, as mentioned earlier, this design presents the potential for harvesting wave energy. The platform's vertical movement in the presence of waves enables the extraction of energy, offering an opportunity to conserve energy by leveraging the heave motion.

\section{Outlook}
In conclusion, our study introduces an innovative hydrofoil propulsion system inspired by aquatic organisms' oscillatory motion. Through simulations, we demonstrated that the camber-modulating hydrofoil significantly enhances propulsive force, providing a 40\% increase in horizontal thrust compared to symmetric alternatives. A prototype flipper, with individual control of heave, pitch, and camber, was manufactured to demonstrate the propulsion system's real-world viability. 
The incorporation of the cambering system significantly enhances the system's capabilities by adding an extra level of maneuverability for steering, along with increased thrust force production.
Looking ahead, further refinements and optimizations, coupled with real-world testing, will validate the system's performance. The redundancy in control mechanisms offers opportunities for adaptive strategies, enhancing maneuverability. This hydrofoil system not only advances underwater robotics but also contributes to sustainable and energy-efficient propulsion technologies. Future research may explore interdisciplinary collaborations to propel advancements in underwater exploration and environmental monitoring.

\bibliographystyle{IEEEtran}
\bibliography{references}

\end{document}